\PassOptionsToPackage{dvipsnames,table}{xcolor}
\documentclass{l4dc2027}
\usepackage[T1]{fontenc}
\usepackage{times,microtype}
\usepackage{mathtools,bm,booktabs,array,ragged2e,tikz,wrapfig,capt-of}
\usepackage[nameinlink]{cleveref}
\hypersetup{colorlinks=true,linkcolor=blue!55!black,citecolor=blue!55!black,urlcolor=blue!55!black}
\definecolor{oursmint}{HTML}{E2F1EA}
\definecolor{losssalmon}{HTML}{F9E8E4}
\definecolor{controlblue}{HTML}{0D47A1}
\definecolor{baselinered}{HTML}{BF360C}
\newtheorem{problem}{Problem}
\newtheorem{assumption}{Assumption}
\crefname{section}{Sec.}{Secs.}
\Crefname{section}{Sec.}{Secs.}
\crefname{subsection}{Sec.}{Secs.}
\Crefname{subsection}{Sec.}{Secs.}
\crefname{subsubsection}{Sec.}{Secs.}
\Crefname{subsubsection}{Sec.}{Secs.}
\crefname{figure}{Fig.}{Figs.}
\Crefname{figure}{Fig.}{Figs.}
\crefname{table}{Tab.}{Tabs.}
\Crefname{table}{Tab.}{Tabs.}
\crefname{algorithm}{Alg.}{Algs.}
\Crefname{algorithm}{Alg.}{Algs.}
\crefname{definition}{Def.}{Defs.}
\Crefname{definition}{Def.}{Defs.}
\crefname{lemma}{Lem.}{Lems.}
\Crefname{lemma}{Lem.}{Lems.}
\crefname{proposition}{Prop.}{Props.}
\Crefname{proposition}{Prop.}{Props.}
\crefname{theorem}{Thm.}{Thms.}
\Crefname{theorem}{Thm.}{Thms.}
\crefname{corollary}{Cor.}{Cors.}
\Crefname{corollary}{Cor.}{Cors.}
\crefname{assumption}{Assum.}{Assums.}
\Crefname{assumption}{Assum.}{Assums.}
\crefname{remark}{Rem.}{Rems.}
\Crefname{remark}{Rem.}{Rems.}
\crefname{problem}{Prob.}{Probs.}
\Crefname{problem}{Prob.}{Probs.}
\crefalias{algocf}{algorithm}
\newcommand{\R}{\mathbb R}
\newcommand{\E}{\mathbb E}
\newcommand{\Prob}{\mathbb P}

\newcommand{\U}{\mathcal U}
\newcommand{\ind}{\mathbf 1}

\newcommand{\pos}[1]{[#1]_+}

\title[Residual Learning Allocation for Multi-Phase Tasks]{Improving Imitation Policies for Multi-Phase Tasks through Residual Learning Allocation}
\coltauthor{\Name{Clinton Enwerem} \Email{enwerem@umd.edu}\\
\Name{Saleh Kemal} \Email{skemal@umd.edu}\\
\Name{John S. Baras} \Email{baras@umd.edu}\\
\Name{Calin Belta} \Email{calin@umd.edu}\\
\addr Institute for Systems Research, University of Maryland, College Park, MD, USA
}

\begin{document}
\raggedbottom
\hypersetup{pdftitle={Improving Imitation Policies for Multi-Phase Tasks through Residual Learning Allocation},pdfauthor={Clinton Enwerem, Saleh Kemal, John S. Baras, Calin Belta}}
\maketitle
\thispagestyle{plain}
\begin{abstract}
Residual reinforcement learning corrects an imitation-trained base policy using a limited budget of environment interaction. In a task with several phases, this budget must be divided among the phases, and the phase in which the base policy fails is not necessarily the phase in which a correction most improves task completion. We formalize this distinction through the \emph{completion gain}: the change in full-task completion probability when, during a single phase, a state-feedback controller replaces selected components of the base policy's action. We decompose the completion gain into the probability of reaching the phase and the conditional improvement in completion from the start of the phase, and we estimate it from paired episodes. The positive estimates determine which action components the residual policy corrects and how training episodes are distributed across phases. In stochastic point-mass docking experiments, most failures occurred in the final docking phase, whereas the largest completion gain occurred in the preceding channel phase; gain-based allocation improved completion over allocation proportional to failure counts and, in a variant with a longer approach, over uniform allocation at half of the budget. In simulated bimanual robotic assembly, the completion-gain estimates were positive only for insertion and release, although few failures occurred during release; in a second assembly configuration, a residual policy trained in the release phase improved completion from held-out states at the start of release by 16.9\%.
\end{abstract}


\section{Introduction}\label{sec:introduction}
Residual reinforcement learning (RL) improves a policy trained by imitation without retraining it: the imitation-trained \emph{base policy} is typically fixed, and a \emph{residual policy} learns a bounded correction to its actions from additional interaction with the environment~\citep{ankile2024resip,ankile2025resfit}. The interaction available for this second stage is limited. For instance, many robotic manipulation tasks typically consist of multiple, ordered phases, such as grasping, transporting, inserting, and releasing a part~\cite{liu2026foresight, lei2025rl100, enweremDoesImitationLearningPreserve2026}. For such tasks, the designer must decide in which phases the residual policy is trained and which components of the action, for example which joints, it corrects. We study how to make this decision under a fixed interaction budget.

\begin{figure}[t]
\centering
\begingroup
\sffamily\fontsize{8}{9}\selectfont
\hspace*{.035\linewidth}\begin{minipage}[b]{.415\linewidth}\centering
\includegraphics[width=\linewidth]{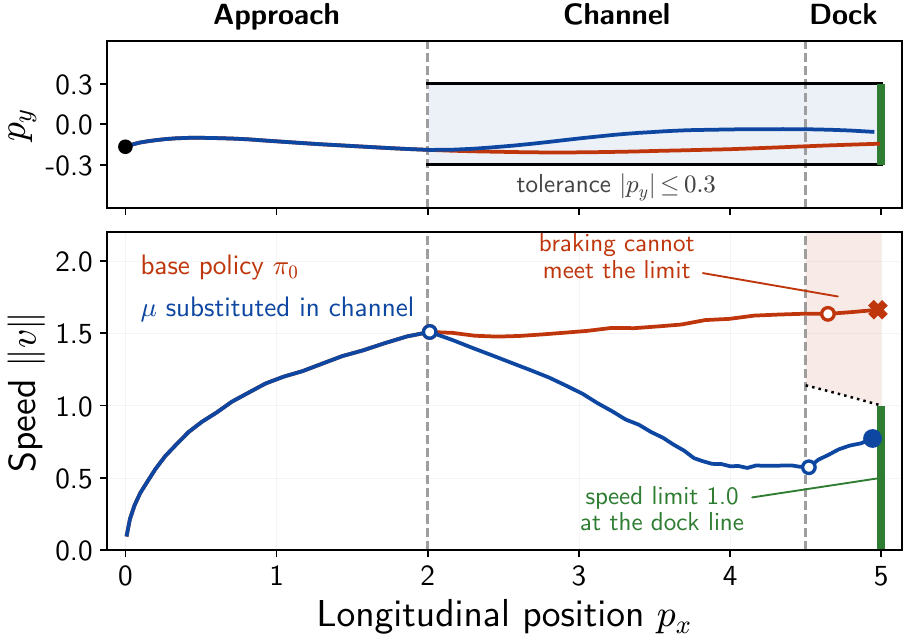}\\[2pt]
{\bfseries (a) Point-Mass Docking}
\end{minipage}\hfill
\begin{minipage}[b]{.478\linewidth}\centering
\begin{tikzpicture}[every node/.style={inner sep=1.5pt}]
\node[inner sep=0pt,anchor=south west] (img) at (0,0) {\includegraphics[width=\linewidth]{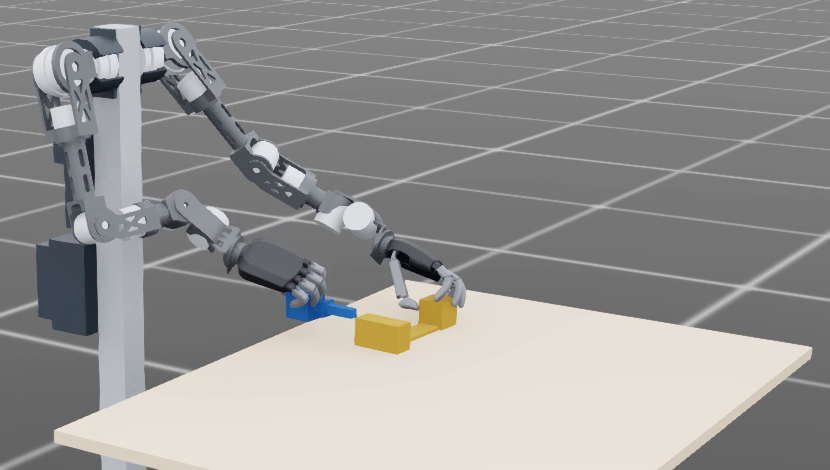}};
\begin{scope}[x={(img.south east)},y={(img.north west)},
  lab/.style={fill=white,fill opacity=.92,text opacity=1,rounded corners=1pt,inner sep=2pt},
  lead/.style={line width=.7pt,black}]
  \node[lab] (rh) at (.19,.15) {Right arm and hand};
  \draw[lead] (rh.north) -- (.335,.435);
  \node[lab] (lh) at (.77,.74) {Left arm and hand};
  \draw[lead] (lh.south west) -- (.545,.41);
  \node[lab] (pt) at (.44,.06) {Part};
  \draw[lead] (pt.north) -- (.385,.335);
  \node[lab] (hs) at (.64,.12) {Housing};
  \draw[lead] (hs.north west) -- (.495,.275);
\end{scope}
\end{tikzpicture}\\[2pt]
{\bfseries (b) Bimanual Assembly}
\end{minipage}\hspace*{.035\linewidth}\\[4pt]
\begin{tikzpicture}[x=.2\linewidth,y=1cm,every node/.style={inner sep=0pt,align=center}]
  \foreach \i/\view/\phase/\cond in {0/grasp/Grasp/reset,1/transport/Transport/part lifted,
      2/insert/Insert/part at opening,3/release/Release/part seated,4/withdraw/Withdraw/hand contact cleared} {
    \node (f\i) at (\i,0) {\includegraphics[width=.192\linewidth]{figures/isaaclab/phase_\view.png}};
    \node[anchor=north] at (\i,-.82) {{\bfseries \phase}\\[-1pt]\cond};
  }
  \foreach \i in {1,2,3,4} {\draw[-latex,line width=1.1pt,white] (\i-.56,0) -- (\i-.44,0);}
\end{tikzpicture}\\[2pt]
{\bfseries (c) Assembly Phases and Initialization Conditions}
\endgroup
\caption{\textbf{Running Examples.} (a) \emph{Point-Mass Docking:} Here, a point mass must pass through a channel and cross the dock line within speed and lateral position limits. The line plots in (a) depict traces from trial runs under two different control settings: a base policy (red) and a control scheme where control of the point-mass's acceleration is switched to a state-feedback controller in the channel phase (blue). In each setting, we use the same initial state, physical parameters, and noise draws. Under the base policy, the point mass exceeds the speed limit at the dock line. With state-feedback acceleration commands in the channel (blue), it satisfies both terminal limits. Open circles mark phase starts. The dotted curve gives the speed bound for one-dimensional braking without drag or noise. (b,c) \emph{Bimanual Robotic Assembly:} In this task, we require a bimanual robotic manipulator to insert a key-like part with its right hand into a housing stabilized by the left hand. The part must remain stable in the housing after the robot withdraws its hands. The frames in the figure show an episode under the state-feedback controller in Isaac Lab~\citep{mittal2023orbit}.}
\label{fig:phase-allocation-concept}
\end{figure}
A natural choice is to train the residual policy in the phases where the base policy fails. However, the phase in which a failure is detected identifies where the episode ended, and not necessarily where a different action would have prevented the failure, because an error made in an early phase can leave the system in a state from which the later phases cannot be completed. \Cref{fig:phase-allocation-concept}(a) shows an example. A point mass with limited braking must pass through a channel and cross a dock line below a speed limit. Under the base policy, the mass enters the dock phase too fast for braking to reduce its speed to the limit, and the failure is detected at the dock line; when a state-feedback controller replaces the base policy during the channel phase only, the episode is completed. Similar behavior can appear in higher-dimensional systems such as the bimanual robotic manipulator depicted in \Cref{fig:phase-allocation-concept}(b,c) charged with contact-rich tasks with multiple phases. In this example, substituting the state-feedback controller during release improved task completion. 

\label{sec:related}\label{sec:related-posttraining}\label{sec:related-phases}
Existing work on improving imitation policies falls into two groups. The first group specifies how the policy is updated. RL can start from demonstrations~\citep{rajeswaran2018dapg,ball2023rlpd} or use human corrections~\citep{luo2025hilserl}. Residual RL trains corrections to a frozen behavior-cloning policy using on-policy~\citep{ankile2024resip} or off-policy updates~\citep{ankile2025resfit}, following earlier residual corrections to conventional controllers~\citep{johannink2019}; other methods steer the denoising inputs of a diffusion policy~\citep{wagenmaker2025dsrl} or use other policy updates~\citep{pi2025recap,lei2025rl100,wang2026lwd}. These methods do not address how limited interaction should be divided among the phases of a task. The second group accounts for the phase structure. Sequential Dexterity improves the states passed between sub-policies~\citep{chen2023sequential}, Foresight residual RL shapes local rewards using predictions of downstream success~\citep{liu2026foresight}, and RL Token and PARTS restrict learning to selected phases~\citep{xu2026rltoken,su2026parts}, which PARTS obtains from bottlenecks identified by a human. To our knowledge, none of these methods estimates, before training, how much a correction restricted to one phase and one group of action components would change full-task completion, or uses such an estimate to divide an interaction budget.

\label{sec:contributions}
In this paper, we estimate the effect of a correction in each phase on completion of the entire task before residual training begins, using the state-feedback controller that generated the demonstrations for the base policy. We compare two episodes that start from the same initial condition (\cref{fig:paired-episodes}): in the first, the base policy supplies every action; in the second, the state-feedback controller replaces selected action components during one phase. We call the resulting change in full-task completion probability the \emph{completion gain}. Our contributions are as follows. First, we define the completion gain and show that it factors into the probability of reaching the phase and the mean improvement in completion probability from the start of the phase. Unlike the failure frequency of a phase, it credits a phase with the failures that a correction in the phase prevents in later phases. Second, we propose \emph{gain-based allocation}, which uses paired estimates of completion gain to select the action components that the residual policy corrects in each phase and the probability with which training episodes start in each phase, with the cost of estimation charged to the same interaction budget. Compared with Foresight~\citep{liu2026foresight}, which changes the reward of the RL update, and PARTS~\citep{su2026parts}, which relies on a human to choose the phases, gain-based allocation leaves the RL update unchanged and chooses the phases from data, using the state-feedback controller. Third, we evaluate the approach in simulated point-mass docking, against four other allocation strategies under the same budget, and in simulated bimanual assembly.

\section{Problem Formulation}\label{sec:formulation}
\subsection{System, Task, and Controllers}\label{sec:state}
We consider a controlled physical system, such as a robot, that must complete a task consisting of ordered phases. In the assembly task of \cref{fig:phase-allocation-concept}(c), for example, the phases are grasp, transport, insert, release, and withdraw. A \emph{task monitor} evaluates conditions on the physical state, such as object poses and contacts, to determine when each phase begins and whether the task has been completed or has failed. The monitor is used for evaluation and for training; its output is not available to the base policy. Three controllers act on the system. A \emph{state-feedback controller} \(\mu\) has access to the full state, completes the task reliably, and supplies demonstrations. The \emph{base policy} \(\pi_0\) is trained by imitation of \(\mu\) but receives only sensor observations, and it completes the task less often. A \emph{residual policy} adds a bounded correction to the action of the frozen base policy and is trained by RL; we allocate the interaction used by its training.

Let \(x(t)\in\mathcal X\) be the physical state of the system and let \(m(t)\in\mathcal M\) collect the internal variables of the controllers and of the task monitor, including observation history, pending actions, elapsed time, and the monitor's task-condition flags. We call \(s(t)=(x(t),m(t))\) the augmented state. An \emph{episode} is one attempt at the task: it starts from an initial augmented state \(s_0\), which we call a reset, and ends at completion, failure, or the horizon \(T\). Its trajectory \(\tau\) records the states and actions. Each episode is generated under an episode condition \(\xi=(s_0,\vartheta,\mathcal O)\) drawn from the evaluation distribution \(\mathcal P\), where \(\vartheta\) collects physical parameters and \(\mathcal O\) specifies the task instance. A policy selects an action \(a_k\in\R^{n_a}\) at each decision time \(t_k\), with \(\Delta t_k=t_{k+1}-t_k\). A fixed execution rule, specified for each task in \cref{sec:exps}, converts the actions to the control reference \(r(t)\in\R^{n_a}\) applied between decisions and includes a constraint map \(\Pi\) that enforces input and reference constraints. With \(s_k=s(t_k)\), the closed-loop transition map is
\begin{equation}
 s_{k+1}=F_{\Delta t_k}\!\left(s_k,
 r\big|_{[t_k,t_{k+1})};\vartheta,\nu_k^x\right),
 \label{eq:dynamics}
\end{equation}
where \(F_{\Delta t_k}\) includes the plant dynamics, the low-level control, and the updates of the controller and task-monitor variables, and \(\nu_k^x\) is a process disturbance. The execution rule is the same for all policies that we compare, so the comparisons differ only in the actions \(a_k\).

\label{sec:information}
The base policy \(\pi_0\) is the learned policy whose task completion we seek to improve. It is an output-feedback policy: at each decision, it maps the observation \(o_k=h(x_k,\nu_k^o)\in\mathcal Y\) and its own memory to the action \(a_k\), where \(x_k=x(t_k)\), \(h\) is the observation map, and \(\nu_k^o\) is an observation disturbance. Learning enters the formulation at two points. First, before the allocation problem is posed, we obtain \(\pi_0\) by behavior cloning: we run \(\mu\) on the system and fit \(\pi_0\) to the resulting pairs of observations and controller actions. Second, the residual policy is trained by RL through interaction with the system~\eqref{eq:dynamics} while \(\pi_0\) is frozen (\cref{sec:allocation}); only this stage consumes the interaction budget.

We assume throughout that the state, observation, and episode-parameter spaces are Borel spaces, that the maps \(F_{\Delta t_k}\) and \(h\), the policies, the controller \(\mu\), and the phase predicates are Borel measurable, and that the disturbances and the internal noise of the policies are drawn from fixed stochastic kernels (Assumption~\ref{ass:borel} in Appendix~\ref{app:proofs}). Because \(m(t_k)\) contains all memory on which the policies and the task monitor depend, the augmented state is then a Markov chain given the episode parameters under each policy \(\alpha\) that we consider, and, by~\citet[Prop.~7.28]{bertsekas1978stochastic}, \(\mathcal P\) and \(\alpha\) induce a unique probability measure \(\Prob_\alpha\) on the space of trajectories, with expectation \(\E_\alpha\).

\subsection{Phases, Completion, and the Allocation Problem}\label{sec:phases}
The task has \(n_p\) ordered phases, indexed by \(j\in\{1,\ldots,n_p\}\). Phase 1 begins at reset, and phase \(j>1\) begins when the task monitor's predicate \(g_j(s_k)\) first holds after phase \(j-1\) has started. Let \(k_j\) be the first decision in phase \(j\), with \(k_j=\infty\) if the phase is not reached, and let \(s^{(j)}=s_{k_j}\) be the phase-start state. The completion indicator \(R(\tau)\in\{0,1\}\) equals one when the task monitor declares the task complete, and the full-task completion probability of a policy \(\pi\) is \(J(\pi)=\E_{\xi\sim\mathcal P,\,\pi}[R(\tau)]\). A failure is detected when a failure condition of the task monitor first holds or the horizon expires; for a failed episode, \(f(\tau)\) denotes the phase in progress when the failure is detected, and we call \(\Prob_\pi(f(\tau)=j)\) the failure frequency of phase \(j\). A \emph{candidate} \(u=(j,G)\) consists of a phase \(j\) and an \emph{action group} \(G\subseteq\{1,\ldots,n_a\}\), that is, a set of action components that a correction may modify during phase \(j\). The interaction budget \(B\) is a number of simulated decisions.

\begin{problem}[Allocation of Residual Training under a Budget]\label{prob:allocation}
Given a base policy \(\pi_0\) trained by imitation of a state-feedback controller \(\mu\), an evaluation distribution \(\mathcal P\), a finite candidate set \(\U\), and an interaction budget \(B\), find an action group \(G_j\) and a phase sampling probability \(w_j\) for each phase \(j\), and train a residual policy that corrects the action components \(G_j\) during phase \(j\) in training episodes that start in phase \(j\) with probability \(w_j\), such that the full-task completion probability \(J\) of the corrected policy is maximized. All simulated decisions used for state collection, completion-gain estimation, residual training, and validation count toward \(B\).
\end{problem}
In other words, the designer chooses where the residual policy acts and where its training episodes start, and every simulated decision spent on this choice reduces the interaction left for training. Our approach (\cref{alg:phase-repair}) estimates, for each candidate, the change in completion probability when \(\mu\) replaces the base policy on the action group during the phase (\cref{sec:repair,sec:estimation}), selects the action groups and phase sampling probabilities from the positive estimates, and trains the residual policy with the remaining budget (\cref{sec:allocation}). The approach rests on the premise that the improvement obtained by \(\mu\) in a phase indicates the improvement available to a learned correction in the same phase; we examine this premise in \cref{sec:toy-results} by training residual policies in single phases.

\section{Method}\label{sec:method}
\subsection{Controller Substitution and Completion Gain}\label{sec:repair}
Let \(\pi\) be a base policy; in our experiments, \(\pi=\pi_0\). For a candidate \(u=(j,G)\), we write \(\mu_u\) for the controller \(\mu\) initialized at the start of phase \(j\) and restricted to the action group \(G\). The substituted policy \(\pi\oplus\mu_u\) acts as follows: during phase \(j\), \(\mu_u\) supplies the components in \(G\) and \(\pi\) supplies the others; before and after phase \(j\), \(\pi\) supplies all components.

\begin{definition}[Completion Gain]\label{def:repair}
The completion gain of candidate \(u=(j,G)\) relative to the base policy \(\pi\) is
\begin{equation}
 \Delta_u(\pi)=J(\pi\oplus\mu_u)-J(\pi).
 \label{eq:repair-effect}
\end{equation}
\end{definition}
The completion gain is a difference of probabilities, and we report it in percent. It quantifies the effect of the controller \(\mu\); for a trained residual policy, we refer to the corresponding difference from the base policy as the \emph{completion improvement}. To relate the completion gain to the phases, let \(V_\alpha(s;\vartheta,\mathcal O)=\E_\alpha[R(\tau)\mid s_k=s,\vartheta,\mathcal O]\) be the probability that policy \(\alpha\) completes the task from augmented state \(s\) with episode parameters \((\vartheta,\mathcal O)\). By the Markov property~\eqref{eq:markov} of Appendix~\ref{app:proofs}, and because the augmented state contains the elapsed time and, at the start of phase \(j\), the initial internal variables of \(\mu_u\), this conditional expectation does not depend on \(k\) and is defined for both \(\pi\) and \(\pi\oplus\mu_u\). Let \(D_u=V_{\pi\oplus\mu_u}-V_\pi\).

\begin{proposition}[Decomposition of the Completion Gain]\label{prop:phase-entry}
Let \(K\) be the terminal decision of an episode, and suppose that the stopped trajectory \((\xi,s_0,\ldots,s_{k_j\wedge K})\) has the same distribution under \(\Prob_\pi\) and under \(\Prob_{\pi\oplus\mu_u}\); that is, episodes under the two policies are identically distributed until phase \(j\) begins, including termination before phase \(j\). If \(\Prob_\pi(k_j<\infty)>0\), then
\begin{equation}
 \Delta_u(\pi)=\Prob_\pi(k_j<\infty)\,\E_\pi\!\left[D_u(s^{(j)};\vartheta,\mathcal O)\,\middle|\,k_j<\infty\right].
 \label{eq:repair-decomposition}
\end{equation}
If \(\Prob_\pi(k_j<\infty)=0\), then \(\Delta_u(\pi)=0\).
\end{proposition}
The proof is given in Appendix~\ref{app:proofs}. The first factor in~\eqref{eq:repair-decomposition} is the probability that the base policy reaches phase \(j\), and the second factor is the mean improvement in completion probability from the phase-start state. The failure frequency \(\Prob_\pi(f(\tau)=j)\) does not appear in either factor: \(D_u\) is defined on completion of the entire task, so a correction during phase \(j\) is credited with the failures that it prevents in phases \(j+1,\ldots,n_p\).

\subsection{Paired Estimation of Completion Gains}\label{sec:estimation}
\begin{wrapfigure}[6]{r}{0.47\linewidth}
\centering
\vspace{-10pt}
\begin{tikzpicture}[x=.505cm,y=.8cm,font=\sffamily\scriptsize,
  every node/.style={inner sep=1.5pt,align=center}]
\node[anchor=east] at (1.9,.83) {Base};
\node[anchor=east] at (1.9,.01) {Substituted};
\foreach \a/\b/\txt in {2.0/4.9/Before,4.9/9.9/Phase \(j\),9.9/12.8/After}{
  \node[anchor=south] at ({(\a+\b)/2},1.12) {\txt};
  \draw[fill=controlblue!8,draw=controlblue!55] (\a,.59) rectangle (\b,1.07);
  \node at ({(\a+\b)/2},.83) {\(\pi_0\)};
}
\foreach \a/\b in {2.0/4.9,9.9/12.8}{
  \draw[fill=controlblue!8,draw=controlblue!55] (\a,-.40) rectangle (\b,.42);
  \node at ({(\a+\b)/2},.01) {\(\pi_0\)};
}
\draw[fill=controlblue!8,draw=controlblue!55] (4.9,-.40) rectangle (9.9,.01);
\draw[fill=baselinered!12,draw=baselinered] (4.9,.01) rectangle (9.9,.42);
\node at (7.4,.215) {\(\mu_u\) on \(G\)};
\node at (7.4,-.195) {\(\pi_0\) on the others};
\end{tikzpicture}
\caption{\textbf{Paired Episodes for \(u=(j,G)\).}}
\label{fig:paired-episodes}
\vspace{-8pt}
\end{wrapfigure}
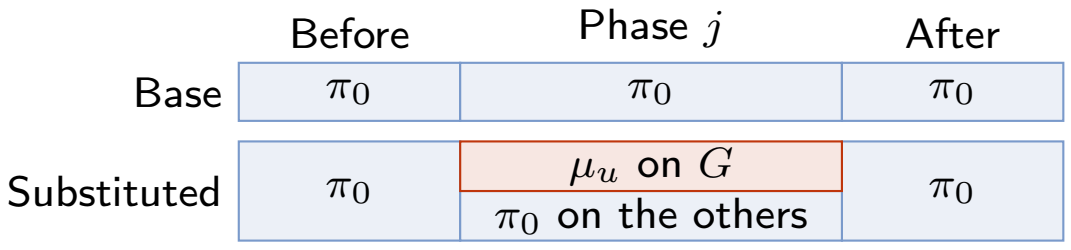
We estimate \(\Delta_u(\pi)\) from \(N\) pairs of episodes (\cref{fig:paired-episodes}). For pair \(i\), we sample an episode condition \(\xi_i\sim\mathcal P\) and run the base policy \(\pi\) and the substituted policy \(\pi\oplus\mu_u\) from this condition with shared policy noise. With completion outcomes \(R_i^\pi\) and \(R_i^u\) and \(d_i=R_i^{u}-R_i^{\pi}\), the estimate and its standard error are
\begin{equation}
 \hat\Delta_u=\frac{1}{N}\sum_{i=1}^{N}d_i,\qquad
 \sigma_u=\Big(\frac{1}{N(N-1)}\sum_{i=1}^{N}\big(d_i-\hat\Delta_u\big)^2\Big)^{1/2}.
 \label{eq:repair-estimate}
\end{equation}
The pairs are independent and \(\E[d_i]=\Delta_u(\pi)\), so \(\hat\Delta_u\) is unbiased and, by the strong law of large numbers, converges almost surely to \(\Delta_u(\pi)\) as \(N\to\infty\). Sharing the episode condition and the policy noise within a pair changes the variance of \(d_i\) but not its mean. Each simulated decision in these episodes counts once toward \(B\); base-policy episodes that are reused for several candidates are counted once.

\subsection{Allocation and Residual Training}\label{sec:allocation}
The allocation makes two choices for each phase: the action group that the residual policy corrects and the probability that a training episode starts in the phase. For phase \(j\), let \(u_j^\star\) be the candidate with the largest estimated gain. We treat candidates whose estimates satisfy \(\hat\Delta_u\ge\hat\Delta_{u_j^\star}-\sigma_{u_j^\star}\) as equivalent to \(u_j^\star\), and among them we select the candidate \(u_j=(j,G_j)\) with the smallest action group, so that the residual policy corrects only the action components that the estimates support. The phase sampling probability of phase \(j\) is
\begin{equation}
 w_j=\frac{\pos{\hat\Delta_{u_j}}}{\sum_{l=1}^{n_p}\pos{\hat\Delta_{u_l}}},
 \label{eq:allocation}
\end{equation}
where \(\pos{\cdot}\) denotes the positive part. Phase \(j\) is allocated if \(w_j>0\); if all selected estimates are nonpositive, the denominator of~\eqref{eq:allocation} is zero, and we do not train a residual policy. Let \(a_k^0\) be the base action supplied by the frozen base policy \(\pi_0\). At decision \(k\) in an allocated phase \(j\), the corrected action is
\begin{equation}
 a_k=a_k^0+\beta\,\Lambda\,\big(\ind_{G_j}\odot\tanh\phi_\theta(z_k)\big),
 \label{eq:residual}
\end{equation}
where \(\phi_\theta:\R^{n_z}\to\R^{n_a}\) is the actor with parameters \(\theta\), \(\ind_{G_j}\in\{0,1\}^{n_a}\) selects the action components in \(G_j\), and \(\odot\) denotes elementwise multiplication. The input \(z_k\in\R^{n_z}\) contains the observation, the normalized base action, and the phase index supplied by the task monitor. The diagonal entries of \(\Lambda\) are the half-ranges of the demonstration actions, bounded below by \(\varepsilon_a>0\), and \(\beta>0\) bounds the correction: \(|a_{k,i}-a^0_{k,i}|<\beta\Lambda_{ii}\) for \(i\in G_j\), and \(a_{k,i}=a^0_{k,i}\) for \(i\notin G_j\) and outside allocated phases.

Training episodes start from phase-start states (\cref{alg:phase-repair}). We first record a set \(\mathcal D_j\) of phase-start states for each phase from episodes of \(\pi_0\). Each training episode then draws a phase \(j\) with probability \(w_j\), restores a state from \(\mathcal D_j\), and applies~\eqref{eq:residual} from the restored state; held-out evaluation always starts from reset, with the residual policy active in every allocated phase. Following~\citet{ankile2025resfit}, we train with off-policy updates, a deterministic actor, a layer-normalized critic ensemble, \(n\)-step returns, and the completion indicator \(R(\tau)\) as terminal reward, until the decision count \(C\) reaches the budget. We assume that every phase has a candidate in \(\U\) and that \(B\) exceeds the cost of collection and estimation.

\begin{algorithm2e}[t]
\footnotesize
\caption{Gain-Based Allocation and Residual Training}\label{alg:phase-repair}
\DontPrintSemicolon
\LinesNumbered
\SetAlgoNoEnd
\KwIn{\(\pi_0,\mu,\mathcal P,\U,N,B\)}
Record phase-start states \(\{\mathcal D_j\}_{j=1}^{n_p}\) under \(\pi_0\); \(C\gets\) decisions used\;
\ForEach{\(u=(j,G)\in\U\)}{
  Estimate \((\hat\Delta_u,\sigma_u)\) from \(N\) paired episodes by~\eqref{eq:repair-estimate}; \(C\gets C+\) new decisions\;
}
\For{\(j=1,\ldots,n_p\)}{
  \(\U_j\gets\{(j,G)\in\U\},\quad u_j^\star\in\arg\max_{u\in\U_j}\hat\Delta_u\)\;
  \(u_j=(j,G_j)\in\arg\min_{\substack{u=(j,G)\in\U_j\\\hat\Delta_u\ge\hat\Delta_{u_j^\star}-\sigma_{u_j^\star}}}|G|\)\;
}
\lIf{\(\sum_j[\hat\Delta_{u_j}]_+=0\)}{stop without training a residual policy}
Set \(w_j\) by~\eqref{eq:allocation} and keep \(\pi_0\) and \(\{G_j,w_j\}\) fixed\;
\While{\(C<B\)}{
  Sample \(j\sim\operatorname{Categorical}(w)\) and restore \(s\sim\mathcal D_j\)\;
  Run the episode from \(s\) with the corrected action~\eqref{eq:residual}; update \(\theta\) off policy with terminal reward \(R(\tau)\)\;
  \(C\gets C+\) new decisions, including validation; save a checkpoint\;
}
\Return{\(\{G_j,w_j\}_{j=1}^{n_p}\) and the residual-policy checkpoints}\;
\end{algorithm2e}

\section{Experimental Setup}\label{sec:exps}
The experiments address two questions: (Q1) does the completion gain identify phases different from those in which failures are detected, and (Q2) does allocating residual training by completion gain improve task completion under a fixed interaction budget? We study these questions in stochastic point-mass docking, simulated in PyTorch (\cref{sec:toy}), and in image-based bimanual assembly, simulated in Isaac Lab~\citep{mittal2023orbit} using PhysX (\cref{sec:protocol}). We address Q1 in both tasks and Q2 in the point-mass task, where we compare gain-based allocation (\cref{alg:phase-repair}) with four strategies that use the same learner; in the assembly task, we train a residual policy in the phase selected by gain-based allocation. \emph{Failure-weighted allocation} sets the phase sampling probabilities proportional to the number of failures detected in each phase; \emph{downstream-failure allocation} weights each phase except the last by the fraction of episodes that reach the next phase and subsequently fail; \emph{uniform allocation} assigns equal probabilities to all phases; and \emph{full-episode training} starts every training episode from reset and corrects every action component in every phase. We also train in a single phase; dock-only training uses the phase with the most detected failures. Every strategy is charged for the base-policy episodes that it uses; gain-based allocation is also charged for gain estimation, and we charge uniform allocation the same estimation cost.

\subsection{Point-Mass Docking}\label{sec:toy}\label{sec:toy-system}
The first case study is a planar point mass that must pass through a channel and cross a dock line below a speed limit (\cref{fig:phase-allocation-concept}(a)). Braking is weaker than forward acceleration, so a failure detected at docking can originate in an earlier phase. The point has unit mass, position \(p=(p_x,p_y)\), velocity \(v\), and state \(x=(p,v)\in\R^4\). The action is an acceleration command \(a_k=(a_x,a_y)\in\R^2\), held constant between decisions. Lengths, speeds, and accelerations are nondimensional, \(\Delta t_k=0.1\)~s, and \(T=16\)~s (160 decisions). The constraint map \(\Pi\) limits the longitudinal acceleration to \([-0.3,1.0]\) and the lateral acceleration to \([-1.0,1.0]\). With drag \(c\), crosswind \(\omega\), and independent noise \(\epsilon_k\sim\mathcal N(0,I_2)\), the dynamics are
\begin{equation}
 v_{k+1}=v_k+\Delta t_k\big(\Pi(a_k)+(0,\omega)-c\,v_k\big)+0.02\sqrt{\Delta t_k}\,\epsilon_k,\qquad
 p_{k+1}=p_k+\Delta t_k\,v_{k+1}.
 \label{eq:toy-dynamics}
\end{equation}
The observation is \(o_k=p_k+0.02\,\eta_k\), with \(\eta_k\sim\mathcal N(0,I_2)\). The base policy computes its action from \(o_k\) alone: we do not provide the velocity, the phase index, or past observations as inputs. Each episode starts at \(p_0=(0,y_0)\) and \(v_0=0\), with \(y_0\), \(c\), and \(\omega\) sampled independently (\cref{tab:toy-parameters}) and fixed within the episode; the evaluation distribution has lower drag than the demonstration distribution, which increases the speed under the same command.

The task has three phases defined by longitudinal position: approach from reset, channel for \(p_x\ge2.0\), and dock for \(p_x\ge4.5\), with the dock line at \(p_x=5.0\). An episode is completed when the point crosses the dock line within \(T\) at speed \(\|v\|\le1.0\) and lateral displacement \(|p_y|\le0.3\); Appendix~\ref{app:toy} lists the failure conditions. For the dynamics without drag and noise, Lemma~\ref{lem:braking} in Appendix~\ref{app:braking-proof} shows that the speed limit can be met only if the speed at the start of the dock phase is at most \(\sqrt{1.0^2+2(0.3)(0.5)}=\sqrt{1.3}\approx1.14\); the dotted line in \cref{fig:phase-allocation-concept}(a) is this bound along the dock phase. We also consider a \emph{long-approach variant}, in which the approach length is increased from 2.0 to 10.0 and the horizon from 160 to 400 decisions, with the channel and dock geometry unchanged.

\label{sec:toy-policies}
The state-feedback controller \(\mu\) regulates the longitudinal speed to a phase-dependent reference and the lateral position to zero, and the base policy \(\pi_0\) is trained by imitation of its clipped commands (Appendix~\ref{app:toy}). At execution, the base action is \(a_k^0=\pi_0(o_k)+0.05\,\zeta_k\), where \(\zeta_k\sim\mathcal N(0,I_2)\) is independent of the process and observation noise. The candidate set \(\U\) combines the three phases with the action groups \(G\in\{\{1,2\},\{1\},\{2\}\}\), that is, both components, the longitudinal component, and the lateral component, which gives nine candidates. The two episodes of a pair coincide until the substitution begins in phase \(j\), as Prop.~\ref{prop:phase-entry} assumes. Under gain-based allocation, the residual policy corrects the selected action groups; under the other strategies, it corrects both components.
\subsection{Bimanual Assembly}\label{sec:protocol}\label{sec:instance}
Our second case study is a simulated bimanual assembly task derived from recent work 
on belief-informed hybrid control for multi-phase manipulation~\citep{enwerem2026beliefinformedhybrid} (\cref{fig:phase-allocation-concept}(b)). Here, the fixed-base bimanual robotic manipulator comprises a pair of 7-degrees-of-freedom (DoF) arms, each equipped with a 6-DoF multi-fingered hand connected by a common torso. Since there are \(n_q=26\) commanded joints, the action has \(n_a=26\) components. The right hand grasps a key-like part, and the left hand supports a movable housing. The part must be inserted within 3~mm and released, and the hands must withdraw without disturbing the seated part. \Cref{fig:phase-allocation-concept}(c) shows the five phases and the conditions that start them; completion requires both objects at rest and the part seated for 2~s, and the task monitor evaluates these conditions from the object states in the simulator (Appendix~\ref{app:phase-details}). Every 50~ms, the policy outputs joint-position targets, which the execution rule interpolates and constrains to joint-position limits and forearm--torso clearance before a joint impedance controller converts them to torques (Appendix~\ref{app:original-controller}).\label{sec:units}

\label{sec:controller}
The state-feedback controller \(\mu\) tracks a 33~s joint trajectory with object-pose feedback and completed 510 of 512 episodes.\label{sec:base-policy} The base policy \(\pi_0\) is a diffusion policy~\citep{chi2023diffusion} that receives torso and right-wrist images and joint readings. We trained it on demonstrations by \(\mu\) and on DAgger data~\citep{ross2011dagger} (Appendix~\ref{app:original-base-training}); \(\pi_0\) completed 15 of 128 episodes. We estimated the completion gain of substituting \(\mu\) on all joints in each phase from 128 paired episodes per phase. Both episodes of a pair received the actions of the base episode until either entered phase \(j\) or terminated; each then used its own observations, with \(\mu_u\) active during phase \(j\) of the substituted episode. Shared actions in GPU contact simulation do not guarantee the identical distributions before phase \(j\) that Prop.~\ref{prop:phase-entry} assumes; the estimates are the paired differences~\eqref{eq:repair-estimate} under this shared-action protocol.

\label{sec:workcell-study}
To test whether a positive completion gain corresponds to an improvement that a residual policy can learn in the assembly task, we trained a residual policy in a second assembly configuration, which differs in table and camera geometry, object masses, and base-policy inputs (Appendix~\ref{app:workcell-geometry}). In this configuration, release was the only phase with a positive estimated completion gain, 31.3\% [12.5, 56.3] from 16 paired episodes per phase (\cref{tab:workcell-repair}), so gain-based allocation places all training episodes in the release phase. We trained the residual policy on all 26 joints from seven recorded release-start states and then recorded release-start states from base-policy episodes at fourteen held-out seeds; 130 of the 448 episodes reached release. In each of 17 evaluations, we ran 16 paired episodes from these states under shared noise, with and without the residual policy.

\section{Results and Discussion}\label{sec:discussion}\label{sec:results-discussion}
\subsection{Point-Mass Docking}\label{sec:toy-results}
\begin{figure}[tb]
\centering
\small
\includegraphics[width=\linewidth]{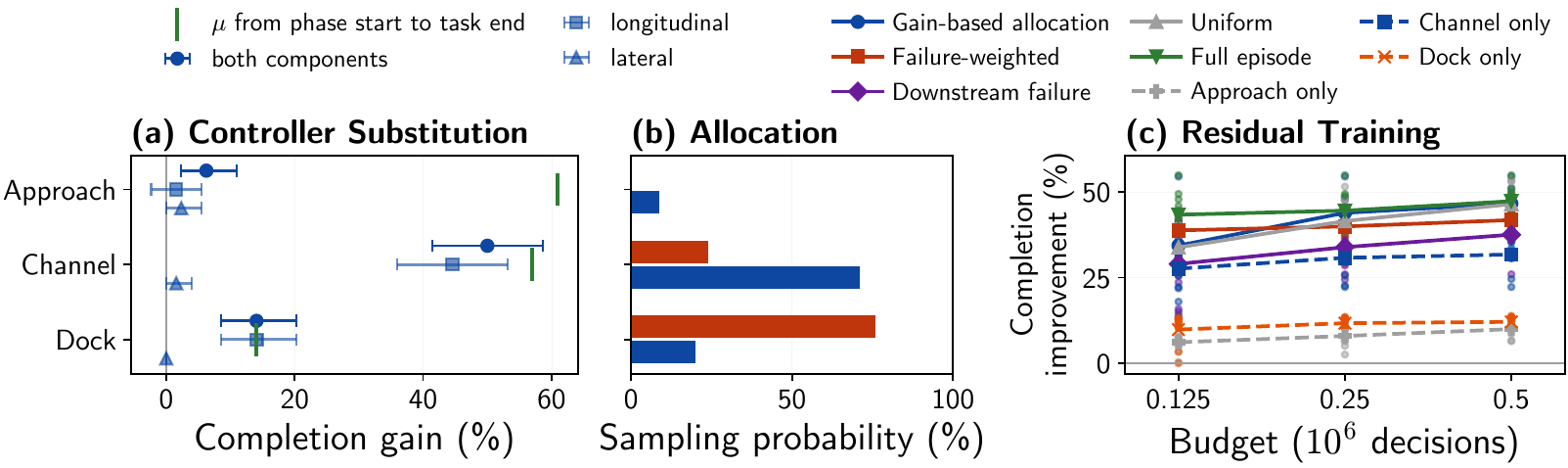}
\caption{\textbf{Completion Gains, Allocation, and Residual Training in Point-Mass Docking.} (a) Estimates from 128 paired episodes per candidate with 95\% bootstrap intervals. (b) Failure-weighted (red) and gain-based (blue) allocation. (c) Means over ten training runs; markers are individual runs, and the budget includes gain estimation.}
\label{fig:toy-repair}
\end{figure}
The completion gain and the failure counts identified different phases. Docking accounted for 101 of the 133 failures in 256 base-policy episodes (\cref{fig:toy-example}(a)), whereas the largest completion gain occurred in the channel phase. In 128 paired episodes, substituting both acceleration components during the channel phase increased completion from 50 to 114 episodes, an estimated completion gain of 50.0\% (95\% paired-bootstrap interval [41.4, 58.6]). The estimated gains for docking and approach were 14.1\% [8.6, 20.3] and 6.25\% [2.3, 10.9], respectively. The selection rule of \cref{sec:allocation} chose both action components for approach and channel and the longitudinal component for docking, and the resulting phase sampling probabilities were 9/71/20\% for approach/channel/dock, compared with 0/24/76\% under failure-weighted allocation (\cref{fig:toy-repair}(a,b)). Consistent with the braking limit of Lemma~\ref{lem:braking}, the state-feedback controller applied from the dock start completed the task only at low dock-start speeds (\cref{fig:toy-example}(c)).

\begin{figure}[t]
\centering
\begin{minipage}[t]{.43\linewidth}
\centering
\captionof{table}{\textbf{Point-Mass Docking at 500,000 Decisions.} Gain-based allocation minus each strategy, in percent, over ten matched training runs. {\setlength{\fboxsep}{1.2pt}\protect\colorbox{oursmint}{Mint}}: interval excludes zero.}
\label{tab:toy-seed-contrasts}
\scriptsize
\setlength{\tabcolsep}{2pt}
\begin{tabular}{@{}lrrr@{}}
\toprule
Strategy & Diff. & Favoring & 95\% interval\\
\midrule
\rowcolor{oursmint}Failure-weighted & +4.8 & 9/10 & [+2.6, +7.2]\\
\rowcolor{oursmint}Downstream & +9.0 & 10/10 & [+5.0, +13.8]\\
\rowcolor{oursmint}Dock only & +34.5 & 10/10 & [+32.0, +37.3]\\
Uniform & +0.1 & 5/10 & [-3.8, +4.0]\\
Full episode & -0.7 & 4/10 & [-4.3, +2.4]\\
\bottomrule
\end{tabular}
\end{minipage}\hfill
\begin{minipage}[t]{.545\linewidth}
\centering
\caption{\textbf{Completion Gains and Allocation in Bimanual Assembly.} Estimates from 128 paired episodes per phase; bars are 95\% bootstrap intervals.}
\label{fig:repair-effects}
\includegraphics[width=\linewidth]{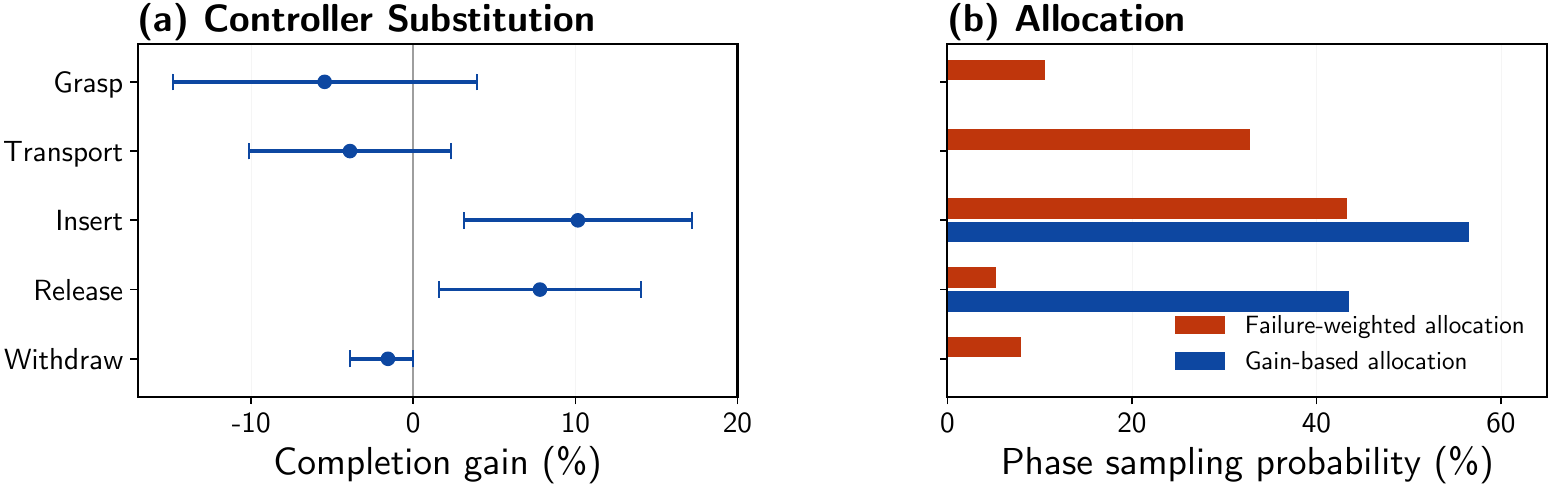}
\end{minipage}
\end{figure}

Residual training followed the same ordering. When the residual policy was trained in a single phase, channel training gave the largest completion improvement, followed by docking and approach (\cref{fig:toy-repair}(c)), consistent with the premise of \cref{sec:phases}. Across ten matched training runs at 500,000 decisions, gain-based allocation achieved a mean completion improvement of 46.6\%, compared with 41.8\% for failure-weighted and 37.5\% for downstream-failure allocation. The paired comparisons in \cref{tab:toy-seed-contrasts} favor gain-based allocation over both of these strategies and over dock-only training, whereas its differences from uniform allocation and full-episode training have intervals that include zero.

In the long-approach variant, 162 of the 164 detected base-policy failures (99\%) occurred at docking. Across ten matched training runs, gain-based allocation achieved a mean completion improvement of 50.1\% at 500,000 decisions, compared with 26.9\% for failure-weighted allocation and 8.4\% for dock-only training (Appendix~\ref{app:toy}, \cref{tab:toy-longapproach}). At 250,000 decisions, gain-based allocation exceeded uniform allocation by 14.1\% [6.7, 21.1] and failure-weighted allocation by 21.3\% [17.6, 25.2]; at 500,000 decisions, its difference from uniform allocation was 0.3\% [-3.4, 4.4]. In this variant, gain-based allocation therefore improved completion over failure-weighted allocation at both budgets and over uniform allocation only at the smaller budget.
\subsection{Bimanual Assembly}\label{sec:results}
\label{sec:repair-results}
In the first assembly configuration, only insertion and release had positive estimated completion gains under substitution on all joints (\cref{fig:repair-effects}(a)): 10.2\% [3.1, 17.2] for insertion and 7.8\% [1.6, 14.1] for release; Appendix~\ref{app:repair-counts} gives the paired counts. The gain-based rule therefore yields phase sampling probabilities of 57\% for insertion and 43\% for release, whereas the failure-weighted rule yields 43\% for grasp and transport together and 5\% for release (\cref{fig:repair-effects}(b)). Controller substitution thus improved completion in release although few failures were detected there.

\label{sec:post-training}
In the second assembly configuration, the residual policy trained in the release phase improved completion from release-start states. Over the 272 paired episodes of the 17 evaluations, the base policy completed the remainder of the task in 128 episodes and the residual policy in 174, a completion improvement of 16.9\% (95\% interval [7.3, 26.5], from a bootstrap over the 130 release-start states). The estimated improvement was positive in fifteen of the 17 evaluations. Because the residual policy acts only from the start of release, the argument of Prop.~\ref{prop:phase-entry} applies to it: its completion improvement from reset is the product of the probability of reaching release and the mean improvement from the release-start state. With 29\% of base-policy episodes reaching release, this product is approximately \(0.29\times16.9\%\approx4.9\%\). In 288 episodes per condition from reset, the base policy completed 50 episodes and the residual policy 47, a difference in completion rate of \(-1.0\%\) [-6.9, 5.2]; this interval contains both zero and the predicted value.

\label{sec:discussion-limits}
In both tasks, the completion gain and the failure counts thus ranked the phases differently. The advantage of gain-based over uniform allocation appeared only at half of the budget in the long-approach variant.
\section{Conclusion}\label{sec:conclusion}
We introduced the completion gain to quantify how substituting a state-feedback controller in one phase changed full-task completion. We showed that under a common trajectory distribution through phase entry, the gain factored into the probability of reaching the phase and the mean change in completion probability from the phase-start state. We used paired gain estimates to select action components for correction and allocate residual training across phases. In simulated point-mass docking, we achieved higher completion rates with gain-based allocation than with failure-weighted allocation under the same nominal decision budget. In the longer-approach variant, we also outperformed uniform allocation at 250,000 decisions. In simulated bimanual assembly, we used controller substitution to select release for residual training. We measured a completion-rate difference of 16.9\% from held-out release-start states and found no clear improvement from reset. We will investigate how training residual corrections in earlier phases affects the probability of reaching later phases and the probability of completing the task conditional on reaching them.

\clearpage
\appendix
\section{Proofs}\label{app:proofs}
\begin{assumption}[Measurability and Noise]\label{ass:borel}
The spaces \(\mathcal X\), \(\mathcal M\), and \(\mathcal Y\) and the spaces of episode parameters are Borel spaces. The maps \(F_{\Delta t_k}\) and \(h\), the policies, the controller \(\mu\), and the phase predicates are Borel measurable. Conditionally on \((s_k,\vartheta,\mathcal O)\), the disturbances \(\nu_k^x\), \(\nu_k^o\) and the internal noise of the policies are drawn from fixed stochastic kernels, independently of the past.
\end{assumption}
The variables \(m(t_k)\) contain all memory on which the policies and the task monitor depend, so under each policy \(\alpha\) that we consider, the action \(a_k\) is a function of \(s_k\), \(o_k\), and the policy noise. Under Assumption~\ref{ass:borel}, the augmented state is therefore a Markov chain given the episode parameters: for every Borel set \(A\subseteq\mathcal X\times\mathcal M\),
\begin{equation}
 \Prob_\alpha\!\left(s_{k+1}\in A\,\middle|\,s_0,\ldots,s_k,\vartheta,\mathcal O\right)=\Prob_\alpha\!\left(s_{k+1}\in A\,\middle|\,s_k,\vartheta,\mathcal O\right).
 \label{eq:markov}
\end{equation}
\subsection{Proof of Proposition~\ref{prop:phase-entry}}\label{app:completion-gain-proof}
\begin{proof}
Let \(\pi^u=\pi\oplus\mu_u\), \(H=\{k_j<\infty\}\), and \(Y=(s^{(j)},\vartheta,\mathcal O)\). Because episodes under the two policies have the same distribution before phase \(j\), we have \(\nu(A)=\Prob_\pi(H,Y\in A)=\Prob_{\pi^u}(H,Y\in A)\) and the same expected completion indicator \(c\) for episodes that terminate before phase \(j\). Conditioning on \(H\) yields \(J(\alpha)=c+\int V_\alpha(y)\,\nu(dy)\) for \(\alpha\in\{\pi,\pi^u\}\). Thus \(\Delta_u(\pi)=\int D_u(y)\,\nu(dy)\). For \(q=\Prob_\pi(H)>0\), dividing by \(q\) gives~\eqref{eq:repair-decomposition}; for \(q=0\), the integral vanishes.
\end{proof}

\subsection{Continuous Braking Condition}\label{app:braking-proof}
To characterize the braking limit of the point-mass task, we consider a continuous, one-dimensional docking segment without drag or process noise. Let \(p\) denote the longitudinal displacement from the start of the segment, \(v\) the forward speed, and \(a\) the acceleration.
\begin{lemma}[Necessary Braking Condition]\label{lem:braking}
Let \(p,v\) be absolutely continuous and \(a\) integrable on \([0,t_c]\), where \(0<t_c<\infty\), with \(\dot p=v\), \(\dot v=a\), and \(a\ge-b\) almost everywhere, where \(b\ge0\). Suppose \(p(0)=0\), \(p(t_c)=\ell>0\), and \(v(t)\ge0\) on this interval. If \(v(t_c)\le v_{\max}\), where \(v_{\max}\ge0\), then
\begin{equation}
 v(0)^2\le v_{\max}^2+2b\ell.
 \label{eq:braking-bound}
\end{equation}
\end{lemma}
\begin{proof}
By absolute continuity, \(\frac{d}{dt}v^2=2va\ge-2bv=-2b\dot p\) almost everywhere. Integrating over \([0,t_c]\) and using \(p(t_c)-p(0)=\ell\), we obtain \(v(t_c)^2-v(0)^2\ge-2b\ell\). Since \(0\le v(t_c)\le v_{\max}\), rearrangement yields~\eqref{eq:braking-bound}.
\end{proof}

\section{Experimental Details}\label{app:experimental-details}
\subsection{Point-Mass Docking}\label{app:toy}
\begin{figure}[tb]
\centering
\includegraphics[width=\linewidth]{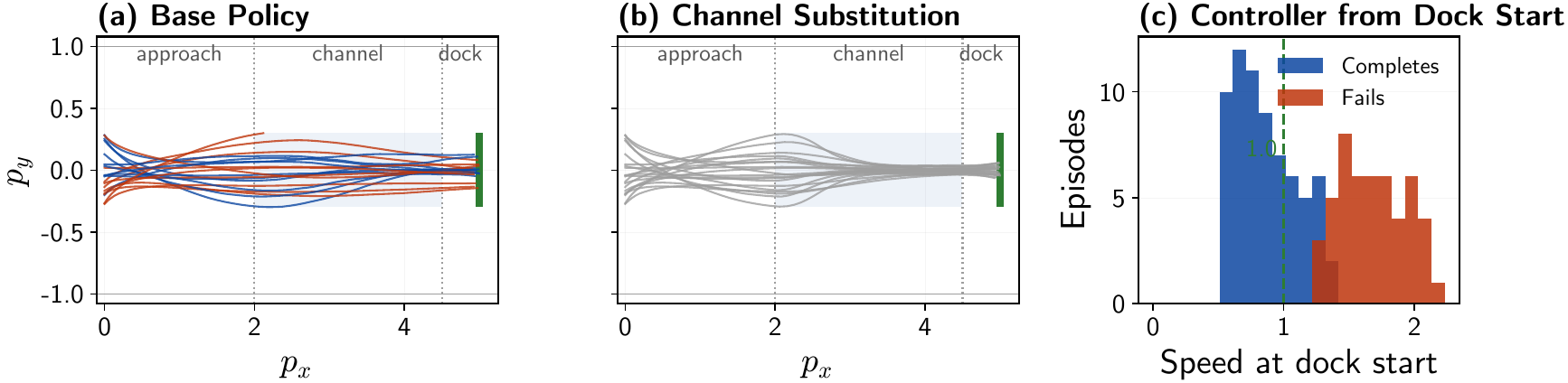}
\caption{\textbf{Point-Mass Trajectories and Speed at the Dock Start.} (a,b) The first 20 of 128 paired episodes: (a) the base policy, with completed episodes in blue and failed episodes in red; (b) the same episode conditions with the state-feedback controller substituted on both action components during the channel phase. Shading marks \(|p_y|\le0.3\) within the channel, and the green bar marks the dock line. (c) Speed at the dock start for the 117 base-policy episodes that reached docking, grouped by whether the state-feedback controller, applied from the dock start, completed the task (68) or failed (49). The dashed line marks the speed limit of 1.0 at the dock line; the bound on the speed at the dock start for the dynamics without drag is 1.14 (\cref{sec:toy-system}).}
\label{fig:toy-example}
\end{figure}
\begin{table}[t]
\centering
\caption{\textbf{Point-Mass Docking Settings.} Lengths, speeds and accelerations are nondimensional.}
\label{tab:toy-parameters}
\footnotesize
\setlength{\tabcolsep}{3pt}
\begin{tabular}{@{}p{0.52\linewidth}>{\raggedleft\arraybackslash}p{0.44\linewidth}@{}}
\toprule
Decision interval \(\Delta t_k\) / horizon \(T\) & 0.1~s / 16~s (160 decisions)\\
Acceleration bounds: longitudinal / lateral & \([-0.3,\,1.0]\) / \([-1.0,\,1.0]\)\\
Gate / channel exit / dock line (\(p_x\)) & 2.0 / 4.5 / 5.0\\
Lateral tolerance / outer lateral bound & 0.3 / 1.0\\
Dock crossing: maximum speed / maximum \(|p_y|\) & 1.0 / 0.3\\
Process noise (velocity, per \(\sqrt{\text{s}}\)) / observation noise & 0.02 / 0.02\\
Demonstration distribution: \(p_y(0)\), drag, crosswind & \(\mathcal U(-0.4, 0.4)\), \(\mathcal U(0.8, 1.2)\), \(\mathcal U(-0.1, 0.1)\)\\
Evaluation distribution \(\mathcal P\): \(p_y(0)\), drag, crosswind & \(\mathcal U(-0.3, 0.3)\), \(\mathcal U(0.05, 0.8)\), \(\mathcal U(-0.15, 0.15)\)\\
Reference speeds (approach, channel, dock) / gains \(k_v,k_p,k_d\) & 1.0, 0.6, 0.6 / 3.0, 4.0, 3.0\\
Demonstrations / base policy / training steps & 200 episodes / MLP 2--64--64--2, tanh output / 4000 Adam steps\\
Base-policy execution noise (std.\ dev.) & 0.05\\
Residual features / actor / bound \(\beta\) & two position observations, \(a_k^0\), phase one-hot / MLP 9--64--64--2, layer norm, tanh / 1.0\\
Critics / hidden widths / target subset & 5 with layer norm / (128, 128) / min over 2\\
Return length / discount / batch size & 5 / 0.995 / 256\\
Updates per stored decision / actor delay / Polyak coefficient & 2 / 2 / 0.005\\
Exploration std.\ dev.\ / target smoothing std.\ dev.\ (clip) & 0.2 / 0.1 (0.3)\\
Adam learning rate, actor / critic & \(3\times10^{-4}\) / \(10^{-3}\)\\
Replay capacity / minimum size for updates & \(5\times10^5\) / 2000\\
Episodes per round / validation every / validation episodes & 64 / 4 rounds / 32\\
Base-policy evaluation episodes / paired episodes per candidate / bootstrap draws & 256 / 128 / 20{,}000\\
Budget \(B\) per strategy / milestones / held-out resets / training runs per strategy & \(500{,}000\) / \(B/4\), \(B/2\), \(B\) / 512 / 10\\
\bottomrule
\end{tabular}
\end{table}
\begin{table}[t]
\centering
\caption{\textbf{Long-Approach Variant at 500,000 Decisions.} Completion improvement over the paired base policy (mean and range over ten matched training runs) and gain-based allocation minus each strategy, in percent. Each run is evaluated on 512 held-out resets. Brackets give paired 95\% intervals; we did not compute an interval for dock-only training. {\setlength{\fboxsep}{1.2pt}\protect\colorbox{oursmint}{Mint}}: the interval excludes zero. Uniform allocation is charged the same gain-estimation cost as gain-based allocation.}
\label{tab:toy-longapproach}
\footnotesize
\setlength{\tabcolsep}{3pt}
\begin{tabular}{@{}lrrr@{}}
\toprule
Strategy & Mean (range) & Difference [95\% interval] & Favoring \\
\midrule
Gain-based & 50.1 (45.9--59.2) & --- & --- \\
Uniform & 49.8 (44.3--54.5) & +0.3 [-3.4, +4.4] & 4/10 \\
Downstream failure & 47.8 (42.0--57.6) & +2.3 [-1.5, +6.0] & 5/10 \\
\rowcolor{oursmint}Full episode & 45.8 (30.1--51.2) & +4.3 [+0.1, +9.5] & 7/10 \\
\rowcolor{oursmint}Failure-weighted & 26.9 (23.0--31.6) & +23.2 [+19.6, +26.7] & 10/10 \\
Dock only & 8.4 (1.2--11.9) & +41.7 & 10/10 \\
\bottomrule
\end{tabular}
\end{table}
The approach, channel, and dock segments have lengths 2.0, 2.5, and 0.5. An episode fails when the crossing of the dock line violates the speed or lateral tolerance, when \(|p_y|>0.3\) within \(2.0\le p_x<5.0\), when \(|p_y|>1.0\) anywhere, or when the horizon expires. The failure phase \(f(\tau)\) is the phase of the position after the failing transition, or the current phase at a timeout. The state-feedback controller is \(a_x=3.0\,(v^{\mathrm{ref}}_j-v_x)\), \(a_y=-4.0\,p_y-3.0\,v_y\), followed by \(\Pi\); its speed references in the three phases are 1.0, 0.6, and 0.6, and we do not provide \(c\) or \(\omega\) to the controller. The same controller supplies the demonstrations and the substitutions. The base policy \(\pi_0\) is an MLP trained on the clipped actions of this controller and noisy position observations from 200 demonstrations (\cref{tab:toy-parameters}). We recorded failures and phase-start states in 256 base-policy episodes and estimated each completion gain from 128 paired episodes with shared episode conditions and noise. The residual policy observes the latest position, a finite-difference velocity estimate, the base action, and the phase indicator. In a training episode, it acts in the sampled start phase and in every later allocated phase; under full-episode training, it acts throughout. The corrected action of the residual policy passes through \(\Pi\), and the learner of \cref{sec:allocation} uses the settings in \cref{tab:toy-parameters}. The last layer of the residual actor is initialized to zero.

In the 128 paired episodes, substituting the longitudinal acceleration component during the channel phase increased completion by 44.5\%, and substituting the lateral acceleration component alone increased completion by at most 2.3\% in any phase. Failure-weighted allocation assigned phase sampling probabilities of 0/24/76\% to approach/channel/dock, downstream-failure allocation assigned 57/43/0\%, and gain-based allocation assigned 9/71/20\% (\cref{fig:toy-repair}(a,b)). To examine the braking limit, we ran the state-feedback controller from the dock-phase start state of each of the 117 base-policy episodes that reached docking. The controller completed the task for all speeds at the dock start below 1.302 and failed for all speeds above 1.3304; the five episodes between these values gave two completions and three failures (\cref{fig:toy-example}(c)). The median speed at the dock start was 1.18, compared with the bound of 1.14 for the dynamics without drag (\cref{sec:toy-system}). \Cref{fig:toy-repair}(a) also marks, for each phase, the gain obtained when the state-feedback controller is applied from the phase start to the end of the task.

The nine sets of paired episodes reused the same 128 base-policy episodes, whose decisions we counted once. Gain-based and uniform allocation each used 86,095 decisions for collection and estimation; the remaining strategies used 13,537 decisions for base-policy evaluation. In the long-approach variant, gain estimation used 170,000 decisions. We trained ten residual policies per strategy with one base policy and with random seeds matched across strategies. We evaluated each trained policy on 512 held-out resets shared across strategies for the corresponding seed; the base-policy and residual-policy episodes of a pair shared the reset and the noise. Intervals for comparisons between strategies use 20,000 bootstrap resamples of paired training runs.

A training episode that starts from a recorded augmented state restores the physical state, decision index, previous position observation, drag coefficient, and crosswind, and uses fresh noise. Training proceeded in rounds of 64 episodes with exploration noise. Each episode started in a phase drawn according to the phase sampling probabilities, from a state recorded at the start of the drawn phase; under full-episode training, each episode started from reset. We stored one \(n\)-step transition per residual decision and assigned the completion indicator \(R(\tau)\) as the only reward at the last residual decision. After 2000 replay transitions had been collected, we performed 2 critic updates per newly stored transition after each round, with actor and Polyak target updates every second critic update. Every 4 rounds, we ran 32 validation episodes from reset without exploration noise and with the residual policy active in all allocated phases, and we retained the checkpoint with the highest validation completion count, replacing it only after a strictly higher count. Training and validation decisions count toward \(B\), and training stopped after the round that exhausted the budget remaining after collection and estimation (\cref{sec:allocation}); across the 80 training runs, the totals ranged from 500,000 to 507,390 decisions. On the held-out episodes, we evaluated the checkpoints retained when the cumulative decisions first reached \(B/4\), \(B/2\), and \(B\); decisions in held-out evaluation do not count toward \(B\). Training and validation used a batched PyTorch implementation of~\eqref{eq:toy-dynamics}, and gain estimation and held-out evaluation used a NumPy implementation.

\subsection{Completion Gains in the Assembly Task}\label{app:repair-counts}
\Cref{tab:repair} reports the paired counts underlying \cref{fig:repair-effects}.
\begin{table}[t]
\centering
\caption{\textbf{Completion Gains in the Assembly Task.} Paired differences in completion, in percent, between the substituted policy (\(\mu\) on all joints during phase \(j\)) and the base policy \(\pi_0\) under the shared-action protocol, from \(N=128\) paired episodes per phase on the same 128 resets. \(N_j\): pairs whose base-policy episode reaches phase \(j\). Gained (lost): pairs completed only by the substituted (base-policy) episode. Failures: episodes with detected-failure phase \(f(\tau)=j\) among the 113 failures in 128 episodes of \(\pi_0\) on the same resets.}
\label{tab:repair}
\footnotesize
\setlength{\tabcolsep}{3pt}
\begin{tabular}{lcccccc}
\toprule
Phase & \(N_j\) & Gained & Lost & \(\hat\Delta_u\) (\%) & 95\% interval & Failures\\
\midrule
Grasp & 128 & 15 & 22 & \(-5.5\) & \([-14.8,\,3.9]\) & 12\\
Transport & 126 & 6 & 11 & \(-3.9\) & \([-10.2,\,2.3]\) & 37\\
Insert & 73 & 18 & 5 & \(\phantom{-}10.2\) & \([\phantom{-}3.1,\,17.2]\) & 49\\
Release & 48 & 14 & 4 & \(\phantom{-}7.8\) & \([\phantom{-}1.6,\,14.1]\) & 6\\
Withdraw & 28 & 0 & 2 & \(-1.6\) & \([-3.9,\,0.0]\) & 9\\
\bottomrule
\end{tabular}
\end{table}
Under the shared-action protocol, the estimated difference between the insertion and release completion gains was 2.3\% (95\% interval \([-7.0,\,11.7]\)).

\subsection{State-Feedback Controller and Execution Rule}\label{app:original-controller}
\label{sec:related-example}
Each arm has seven joints, and each hand has six actuated and five passive joints, which gives 36 joints, of which 26 are commanded. In the assembly task, the execution rule interpolates consecutive joint-position targets and applies the constraint map \(\Pi\):
\begin{equation}
 r(t)=\Pi\!\left(a_{k-1}+\tfrac{t-t_k}{\Delta t_k}\,(a_k-a_{k-1})\right),\qquad t\in[t_k,t_{k+1}),
 \label{eq:issued-reference}
\end{equation}
where \(\Pi\) enforces joint-position limits and a 2~mm forearm--torso clearance margin. A joint impedance controller with Coriolis and gravity compensation converts \(r(t)\) to torques. We integrated the dynamics at 2~ms, so each 50~ms policy decision spans 25 integration steps, over a horizon of \(T=75\)~s, and we randomized the initial object positions within squares of 5~mm half-width.

The state-feedback controller \(\mu\), which supplies the demonstrations and the substitutions \(\mu_u\), tracks a 33~s joint reference parameterized by \(\varphi\in[0,1]\). The rate of \(\varphi\) decreases with the mechanical energy of the robot and the objects and with the reference-tracking error, and conditions on the relative part--housing pose hold \(\varphi\) at the transfer, entry, and seating events of the reference trajectory. During transport and insertion, the controller corrects the right-arm reference using right-palm pose feedback to track the housing-relative part target, subject to joint-limit and displacement bounds in a least-squares solve. At the start of release, it resets the right-arm reference to the current joint configuration to remove insertion preload before opening the right hand. The left hand stabilizes the housing until the part clears hand contact as the right hand withdraws; the left-hand joint reference then relaxes to the current joint configuration. During opening, the controller bounds each finger reference offset from the current joint position before imposing the final joint bounds and thumb-reference lower bounds; the left arm then withdraws. \Cref{tab:parameters} lists the durations and the opening offset. With access to the true object poses, which the base policy does not observe, \(\mu\) completed 510 of 512 episodes drawn from \(\mathcal P\) within \(T\).

\subsection{Base-Policy Training}\label{app:original-base-training}
The assembly task uses a 0.35~kg housing and a 0.12~kg part. The observation map is \(h(x_k)=((I_k^v)_{v\in\mathcal V},q_k,\dot q_k)\), where \(I_k^v\) is the RGB image from camera \(v\) and \(q_k,\dot q_k\in\R^{n_q}\) are the joint position and velocity readings. The policy does not observe object poses, contact state, or the internal state of the state-feedback controller. The cameras are a torso camera and a right-wrist camera, \(\mathcal V=\{\text{torso},\text{wrist}\}\), each providing \(128\times96\) RGB images with intrinsics modeled after a physical stereo RGB-D camera. The torso camera views the workspace from the robot base; the wrist camera views the held part and the housing opening during alignment and insertion. The base policy \(\pi_0\) is a diffusion policy~\citep{chi2023diffusion} conditioned on the latest \(L=2\) observations \(o_k\). It predicts chunks of actions and executes a prefix before replanning (\cref{tab:parameters}). Visual imitation has previously been used for assembly~\citep{zhao2023act,huang2025vtrefine}, and ENPIRE addresses complementary reset and verification infrastructure~\citep{xiao2026enpire}. We trained it on 512 demonstration trajectories of the state-feedback controller (Appendix~\ref{app:original-controller}), with each target action equal to the controller's reference at the end of the corresponding interval. We augmented these data with three rounds of DAgger~\citep{ross2011dagger}, in which the state-feedback controller resumed from randomly selected states visited by earlier learned policies and completed the episode from the selected state. We fixed the probability of sampling these DAgger segments during training; \cref{tab:parameters} reports the interval of the times at which the controller resumed, the sampling probability, and optimization settings. On 128 evaluation resets from \(\mathcal P\), \(\pi_0\) completed 15 episodes.

\begin{table}[t]
\centering
\caption{\textbf{Base-Policy and State-Feedback Controller Parameters for the Assembly Task.}}
\label{tab:parameters}
\footnotesize
\begin{tabular}{@{}lr@{}}
\toprule
\multicolumn{2}{@{}l}{\emph{Release sequence of the state-feedback controller}}\\
Contact-free time of the part before left-hand release & 0.25~s\\
Latest left-hand release after right-hand withdrawal & 1~s\\
Left-hand reference set to current joint positions & 1~s\\
Left-hand opening & 5~s\\
Opening offset before final reference bounds & 0.004~rad\\
Left-arm withdrawal & 2~s\\
\midrule
\multicolumn{2}{@{}l}{\emph{Base policy \(\pi_0\)}}\\
Observation history \(L\) & 2\\
Predicted / executed actions per chunk & 16 / 8\\
DAgger controller-resumption time, uniform on & [12, 50]~s\\
Probability of sampling DAgger segments & 40\%\\
Gradient steps / batch size & \(10^5\) / 64\\
Exponential moving average, maximum decay & 0.9999\\
\midrule
\multicolumn{2}{@{}l}{\emph{Controller substitutions}}\\
Action blending duration for \(\mu_u\) & 0.5~s\\
Pairs \(N\) & 128\\
Bootstrap resamples & 20{,}000\\
\bottomrule
\end{tabular}
\end{table}
\subsection{Phase Predicates and Controller Substitution}\label{app:phase-details}
Completion requires that the part remain within 3~mm laterally, 2~mm vertically, and 0.06~rad of its seated pose in the housing for 2~s. Both objects must be nearly at rest, both hands at least 5~mm from each object, and the table must support at least half their combined weight. The five phases (\(n_p=5\)) begin as follows. Grasp begins at reset. Transport begins when the right hand stabilizes the part with at least two contacts and lifts it from the table. Insertion begins when the part reaches the housing opening with lateral and vertical offsets below 3~mm relative to the housing axis. Release begins upon seating within the terminal tolerances. Withdrawal begins once neither hand exerts more than 0.01~N on either object after seating. At the beginning of phase \(j\), we initialize \(\mu_u\) from the current state at the corresponding stage of its trajectory and blend its action from the base policy's last action, retaining the impedance preload of a held grasp. We evaluated substitutions on all joints, \(G=\{1,\ldots,n_q\}\), with \(N=128\) paired episodes for each phase on the same 128 resets, and we report nonparametric bootstrap intervals across pairs.

\FloatBarrier
\subsection{Second Assembly Configuration and Release-Phase Residual Training}\label{app:workcell-geometry}\label{app:workcell-posttraining}
In the second assembly configuration, the tabletop is at \(z=-0.472\)~m in the shoulder-midpoint frame, and the housing and part start at planar coordinates \((0.490,0.111)\)~m and \((0.450,-0.189)\)~m. Their masses are 105.8~g and 47.4~g, respectively. We scaled inertias with mass at fixed geometry and center of mass, assuming uniform density. GPU PhysX used 64 position and 8 velocity solver iterations and integrated external forces at each iteration. The task uses the completion predicates, 75~s horizon, and reset perturbation range of \cref{sec:instance}. The base policy receives one \(320\times180\) image and 26 joint positions; we trained it for \(10^5\) updates with batch size 16 on 23,751 decisions from 29 completed episodes of the state-feedback controller. It predicts 16 actions and executes eight per chunk. \Cref{tab:workcell-repair} reports the paired completion counts from which we estimated the completion gains in this configuration.
\begin{table}[t]
\centering
\caption{\textbf{Completion Gains in the Second Assembly Configuration.} Completed episodes among 16 paired episodes per phase under the base policy and under the substituted policy; estimated completion gains and 95\% bootstrap intervals are in percent.}
\label{tab:workcell-repair}
\small
\setlength{\tabcolsep}{3pt}
\begin{tabular}{@{}lrrrr@{}}
\toprule
Phase & Base & Substituted & \(\hat\Delta_u\) (\%) & 95\% interval\\
\midrule
Grasp & 2 & 1 & \(-6.3\) & \([-18.8,\,0.0]\)\\
Transport & 6 & 3 & \(-18.8\) & \([-37.5,\,0.0]\)\\
Insert & 3 & 1 & \(-12.5\) & \([-31.3,\,0.0]\)\\
\rowcolor{oursmint} Release & 1 & 6 & \(\mathbf{31.3}\) & \([12.5,\,56.3]\)\\
Withdraw & 2 & 0 & \(-12.5\) & \([-31.3,\,0.0]\)\\
\bottomrule
\end{tabular}
\end{table}

The residual policy corrects all 26 joints during the release phase, and the base policy alone controls the other phases. Training episodes start from seven release-start states recorded from base-policy episodes. The learner uses five critics with hidden width 256, 5-step returns, discount 0.995, batch size 256, an offline replay fraction of 0.5, 10,000 critic-only updates before the actor is updated, learning rates of \(10^{-6}\) for the actor and \(10^{-4}\) for the critics, exploration noise with standard deviation 0.01, and target noise with standard deviation 0.05 clipped at 0.15. We saved a checkpoint every 10,000 decisions and selected the checkpoint at 120,000 decisions, which had the highest completion rate over the most recent 200 training episodes (0.870); we made this selection before recording the held-out states. For evaluation, we recorded release-start states from 32 base-policy episodes from reset at each of fourteen seeds that we did not use for training, which gave between 5 and 15 states per seed and 130 states in total. Each evaluation ran 16 paired episodes; when a seed gave fewer than 16 states, states were repeated under distinct noise draws, and three seeds were evaluated under two noise seeds each. Pairs that share a release-start state are not independent, so the reported interval comes from a weighted bootstrap that resamples release-start states; the bootstrap over the 272 pairs gives [10.3, 23.5]. The comparison from reset used 18 evaluations of 16 episodes per condition, 288 episodes per condition in total. In these evaluations, the base-policy actions of the two conditions differ from the first decision, so the episodes of the two conditions are independent samples, and the reported interval is a bootstrap of the difference between their completion rates.

\end{document}